\documentclass[conference]{IEEEtran}
\IEEEoverridecommandlockouts
\usepackage{cite}
\usepackage{amsmath,amssymb,amsfonts}
\usepackage{bbm}
\usepackage{algorithmic}
\usepackage{graphicx}
\usepackage{textcomp}
\usepackage{xcolor}

\usepackage{hyperref}       
\usepackage{url}            
\usepackage{booktabs}       
\usepackage{nicefrac}       
\usepackage{microtype}      
\usepackage{cleveref}       
\usepackage{lipsum}         
\usepackage{graphicx}

\usepackage{subcaption}
\usepackage{multirow}
\usepackage[linesnumbered,ruled,vlined]{algorithm2e}

\def\BibTeX{{\rm B\kern-.05em{\sc i\kern-.025em b}\kern-.08em
    T\kern-.1667em\lower.7ex\hbox{E}\kern-.125emX}}
\begin{document}

\title{Masked Multi-Step Probabilistic Forecasting for Short-to-Mid-Term Electricity Demand*
\thanks{*This work was supported by GE Digital.}
}

\author{\IEEEauthorblockN{Yiwei Fu}
\IEEEauthorblockA{\textit{GE Research} \\
Niskayuna, NY, USA 12309 \\
\texttt{yiwei.fu@ge.com}}
\and
\IEEEauthorblockN{Nurali Virani}
\IEEEauthorblockA{\textit{GE Research} \\
Niskayuna, NY, USA 12309 \\
\texttt{nurali.virani@ge.com}}
\and
\IEEEauthorblockN{Honggang Wang\IEEEauthorrefmark{2}\thanks{\IEEEauthorrefmark{2}Dr. Wang contributed to this work when he was at GE Research.}}
\IEEEauthorblockA{\textit{Upstart Power} \\
Southborough, MA, USA 01772 \\
\texttt{hgwang2010@gmail.com}}
}

\maketitle

\begin{abstract}
Predicting the demand for electricity with uncertainty helps in planning and operation of the grid to provide reliable supply of power to the consumers. Machine learning (ML)-based demand forecasting approaches can be categorized into (1) sample-based approaches, where each forecast is made independently, and (2) time series regression approaches, where some historical load and other feature information is used. When making a short-to-mid-term electricity demand forecast, some future information is available, such as the weather forecast and calendar variables. However, in existing forecasting models this future information is not fully incorporated. To overcome this limitation of existing approaches, we propose Masked Multi-Step Multivariate Probabilistic Forecasting (MMMPF), a novel and general framework to train any neural network model capable of generating a sequence of outputs, that combines both the temporal information from the past and the known information about the future to make probabilistic predictions. Experiments are performed on a real-world dataset for short-to-mid-term electricity demand forecasting for multiple regions and compared with various ML methods. They show that the proposed MMMPF framework outperforms not only sample-based methods but also existing time-series forecasting models with the exact same base models. Models trainded with MMMPF can also generate desired quantiles to capture uncertainty and enable probabilistic planning for grid of the future. 
\end{abstract}


\section{Introduction}~\label{sec:intro}
The paradigm shift towards low carbon and more intermittent renewable power generation, perplexes the load and generation balancing for the power grid planners and operators. Accurately forecasting electricity demand is becoming more critical for effective power system management, dispatch optimization, maintenance scheduling, energy trading, and capacity planning. These forecasting processes can be grouped into four categories based on their horizons~\cite{hong2016probabilistic}: very short term, short term, mid term, and long term load forecasting, and there are many papers on them~\cite{hong2020energy}. Most relevant to this paper are deep learning-based short-to-mid-term forecasting models that provide multi-step outputs. These include convolutional neural networks (CNN) for short-term load forecasting~\cite{deng2019multi}, sequence-to-sequence long short-term memory (LSTM) for single house load forecasting~\cite{masood2022multi}, and hybrid CNN and LSTM for short-term consumption forecasting~\cite{yan2018multi}. However, accurate forecasting is also becoming difficult due to variability in the behind-the-meter residential-scale, community-scale, and utility-scale distributed renewable energy generation that impacts the net demand. Thus, probabilistic forecasting tools are gaining more importance to help operators understand plausible peak net load that they expect to see on their network. In this work, we focus on a novel framework for short-term and mid-term electricity demand forecasting with uncertainty.


The electricity demand forecasting models can be grouped into two categories according to forecasting steps: one-step forecasting, which estimates future demand one step ahead in time, and multi-step forecasting that predicts multiple time steps into the future. This paper addresses the multi-step forecasting problem. 
Time series models for multi-step forecasting can be categorized into two kinds of approaches: recursive methods and direct methods~\cite{lim2021time}. 
Recursive methods typically use an auto-regressive approach for one-step-ahead prediction and produce multi-step forecasts by recursively feeding predictions as inputs into the future time steps.
However, there is often some error at each step, so the recursive structure tends to accumulate large errors over long forecasting horizons.
Alternately, direct methods directly map all available inputs to multi-step forecasts and typically use a sequence-to-sequence (seq2seq) structure.
The disadvantage of this method is that it is harder to train, especially when the forecast horizon is large~\cite{kline2004methods}.

For short-term and mid-term electricity demand forecasting problem, there is often some estimate or known future information available, such as weather forecasts and calendar variables, how to make most use of these future information is the key focus of this paper. Most existing work does not incorporate known future information directly during training. 
Recursive methods use the future information when making iterative forecasts, but they are trained on 1-step predictions only.
Partially inspired by the success of natural language models such as BERT~\cite{devlin_bert_2019}, and to address the gap in incorporating known future information in multi-step forecasting, we propose Masked Multi-Step Multivariate Probabilistic Forecasting (MMMPF) framework. MMMPF is not a model, but a general self-supervised learning task for training all NN time series models (including recurrent NNs, CNNs, and attention-based models) to make multi-step probabilistic forecasts with known future information. MMMPF is a flexible learning framework that improves upon existing methods by taking into account both recent history and known future information.

The contributions of this paper are as follows:
\begin{itemize}
	\item We propose MMMPF, a novel and general framework for training NN-based multi-step forecasting models with known future information. It uses a masking technique that is flexible and can generate forecasts over different horizons. It improves existing methods by combining both recent history and known future information.
	\item MMMPF can incorporate quantile regression loss functions and generate probabilistic forecasts.
	\item The proposed framework is validated by an electricity demand forecasting challenge. MMMPF shows consistently better results for different forecasting horizons on different models in this real-world dataset. 
\end{itemize}

It should be emphasized that the goal of this paper is not to solve one forecasting problem with the best-tuned model, rather, to offer a new learning framework which could be applied to any NN model. The comparisons with existing time series forecasting methods are by no means exhaustive, but they are fair because they use the same base model and hyperparameters. 
It should also be noted that studying how the future information (i.e., weather forecasts) is generated, or evaluating how good it is, are beyond the scope of this paper.
We have focused on developing a general deep learning framework for time series probabilistic modeling that can incorporate future information in making multi-step forecasts when it is available.

\section{Background}~\label{sec:review}

Machine learning-based forecasting models can be categorized into sample-based models and time-series models.
Traditionally, sample-based regression models such as multiple linear regression and hidden Markov model~\cite{cheng2006multistep}, feed-forward neural networks~\cite{kline2004methods}, nearest neighbors~\cite{sorjamaa2007methodology}, decision trees~\cite{yang2009multi}, support vector regression~\cite{bao2014multi}, ensemble of varied length mixture models~\cite{ouyang2018multi} have been used for forecasting.
To achieve multi-step forecasting, one simply run the sample-based model multiple times for each step.

Time series models can be further categorized into recursive methods and direct methods. 
Recursive methods~\cite{li2019enhancing, salinas2020deepar, lim2020recurrent, suradhaniwar2021time} generate multi-step forecasts by recursively feeding 1-step forward predictions into future time steps.
These approaches often lead to error accumulation because a forecast is based on all previous forecasts.
Direct methods~\cite{hauser2017probabilistic, fan2019multi, hauser2018neural, lim2021temporal, dabrowski2020forecastnet, masood2022multi} map all available input directly to all forecasts using sequence-to-sequence models, but they do not use future inputs.

\begin{figure}[t]
	\centering
	\includegraphics[width=\linewidth]{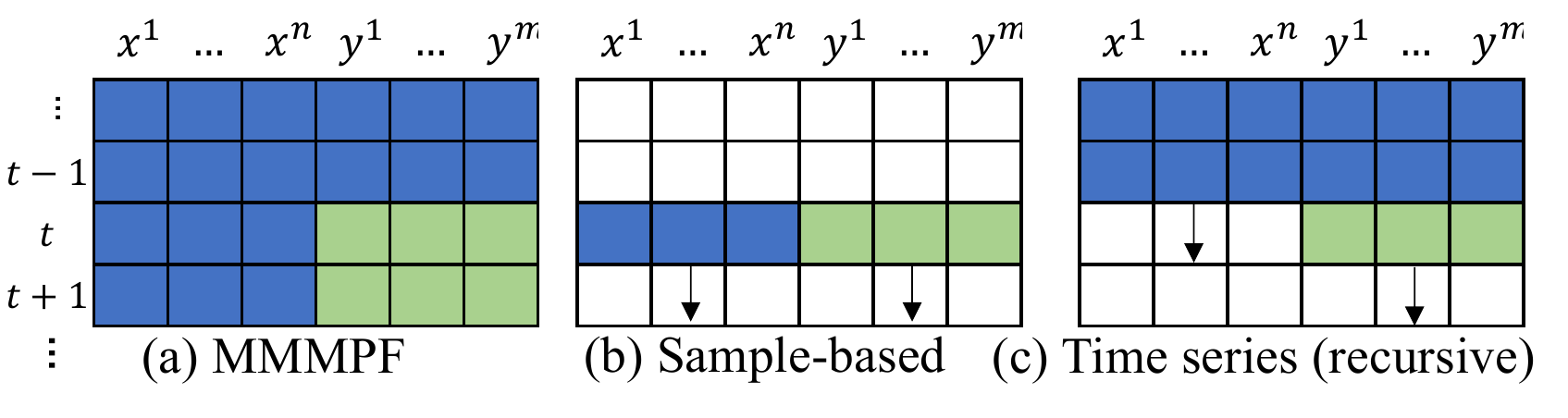}
	\caption{Multi-step forecasting, variables in blue (darker shade) are used to forecast those in green (lighter shade). (a) MMMPF uses all past and future information to predict forecast variables. (b) Sample-based methods do not consider recent history. (c) Existing time series models do not use all future information.}
	\label{fig:formulation}
	\vspace{-0.3cm}
\end{figure}

Figure~\ref{fig:formulation} illustrates the key insight of MMMPF: to make accurate multi-step forecasts, it is imperative to incorporate all past information, as well as any known future information.
Existing methods fall short at some of these aspects: sample-based models do not consider the recent history and time series models do typically use all known future information or make forecasts for multi-steps at once.
In the context of load forecasting, sample-based methods do not consider previous day's load, while time series models do not use all future weather information in their models.
In contrast, MMMPF is designed to maximally take advantage of all available information, past or future, while keeping the forecasting horizon flexible.
This is the intuition why MMMPF could potentially outperform existing forecasting frameworks.

\section{Problem Formulation and Proposed Solution}\label{sec:MMMPF}
\subsection{Masked Multi-Step Multivariate Probabilistic Forecasting}~\label{sec:MMMPF:formulation}
Consider a multivariate time series forecasting problem: let $\mathbf{x_t}\in \mathbb{R}^n$ be a sample of predictor variables with dimension $n$ at time $t$ and the $j$-th dimension is denoted as $x^j_t$ (i.e., $\mathbf{x_t}=[x_t^1, x_t^2,...,x_t^n]$), $\mathbf{y_t}\in \mathbb{R}^m$ be a sample of forecast variables with dimension $m$ at time $t$ (i.e., $\mathbf{y_t}=[y_t^1, y_t^2,...,y_t^m]$), the task is to predict up to $(k+1)$ steps ($k \geq 0$) of forecast variables $\mathbf{y_t}, \mathbf{y_{t+1}}, ..., \mathbf{y_{t+k}}$ from past $T$-step information as well as the future information for predictor variables up to time $t+k$. 
A distinct feature of this problem formulation compared to a standard time series forecasting problem is the need to incorporate future information into the predictions directly. 
For example, when forecasting electric demand for a particular region over the next month, the calendar variables (date, month, day of week, etc.) and some weather forecasts are known, yet traditional forecasting formulations do not take full advantage of the future information.

Our recently proposed Masked Multi-Step Multivariate Forecasting (MMMF)~\cite{fu2022masked} directly models the following:
\begin{equation}~\label{eqn:MMMPF}
		\hat{\mathbf{y}}_t,...,\hat{\mathbf{y}}_{t+k} =  f(\mathbf{x_{t-1}},...,\mathbf{x_{t-T}},\mathbf{y_{t-1}},...,\mathbf{y_{t-T}},\mathbf{x_t},...,\mathbf{x_{t+k}})
\end{equation}
where $f$ is the function being modeled, $\hat{\mathbf{y_t}}$ are estimations of the ground truth $\mathbf{y_t}$ values, $\mathbf{x_{t-1}}, ... ,\mathbf{x_{t-T}}$ are the past predictor variables, $\mathbf{y_{t-1}}, ... ,\mathbf{y_{t-T}}$ are the past forecast variables, and $\mathbf{x_t}, ... ,\mathbf{x_{t+k}}$ are the future predictor variables.
We propose Masked Multi-Step Multivariate Probabilistic Forecasting (MMMPF) by combining MMMPF with quantile regression (QR)~\cite{koenker2001quantile}, i.e., instead of learning a deterministic relationship with mean squared error as the loss function, we learn probabilistic forecasts in MMMPF.
Quantiles provide an interpretable representation for uncertainty because they can be used to model complex distributions without parametric assumptions, and allow for simple construction of prediction intervals.
QR involves optimizing the pinball loss between the ground truth $\mathbf{y}$ and prediction $\hat{\mathbf{y}}$: 
\begin{equation}
    \rho_\tau(\mathbf{y}, \hat{\mathbf{y}}) =  (\hat{\mathbf{y}} - \mathbf{y})(\mathbbm{1}_{\mathbf{y} \leq \hat{\mathbf{y}}} - \tau)
\end{equation}
where $\tau$ is the desired quantile level and $0<\tau<1$ and indicator function $\mathbbm{1}_{\mathbf{y} \leq \hat{\mathbf{y}}} = 1$, if $\mathbf{y} \leq \hat{\mathbf{y}}$, else it is $0$.

In comparison, traditionally there are three most common machine learning formulations for modeling such a multi-step multivariate forecasting problem:
\begin{enumerate}
	\item Sample-based forecasting (SBF) approach: this formulation maps the predictor variables to forecast variables directly without considering the temporal dependency:
	\begin{equation}~\label{eqn:sbf}
		\hat{\mathbf{y}}_{t} = f(\mathbf{x_{t}}), ... ,\hat{\mathbf{y}}_{t+k} = f(\mathbf{x_{t+k}})
	\end{equation}
	\item Recursive single-step forecasting (RSF) approach: this formulation learns a one-step forward prediction model and apply that recursively during inference, i.e.:
	\begin{equation}~\label{eqn:rsf}
		\begin{split}
			\hat{\mathbf{y}}_{t} &= f(\mathbf{x_{t-1}},...,\mathbf{x_{t-T}},\mathbf{y_{t-1}},...,\mathbf{y_{t-T}}) \\ 
			\hat{\mathbf{y}}_{t+1} &= f(\mathbf{x_{t}},...,\mathbf{x_{t-T+1}},\hat{\mathbf{y}}_{t}, \mathbf{y_{t-1}},...,\mathbf{y_{t-T+1}}) \\
		\end{split}
	\end{equation}
	\item Direct multi-step forecasting (DMF) approach: this formulation directly generates multiple steps of forecast variables given past information, i.e.:
	\begin{equation}~\label{eqn:dmf}
		\hat{\mathbf{y}}_t,...,\hat{\mathbf{y}}_{t+k} = f(\mathbf{x_{t-1}},...,\mathbf{x_{t-T}},\mathbf{y_{t-1}},...,\mathbf{y_{t-T}})
	\end{equation}
\end{enumerate}
MMMPF, RSF and DMF are time-series models, while SBF approaches are sample-based.


\subsection{Proposed Algorithm}~\label{sec:MMMPF:solution}
To solve this multi-step multivariate probabilistic forecasting problem, our proposed MMMPF uses a masked time series model (MTSM) approach. 
Inspired by~\cite{fu2022mad} where MTSMs are used for time series anomaly detection and have shown superior performance to traditional regression models, we detail MMMPF training in Algorithm~\ref{algo:MMMPF-train}.
In this algorithm, we replaced all masked variables with random values within the ranges of those variables.
The time series model $f_\theta$ in this algorithm can be any neural network model that generates a sequence of outputs, such as Long Short-Term Memory (LSTM) network~\cite{hochreiter_long_1997}, Temporal Convolutional Network (TCN)~\cite{bai_empirical_2018} Transformer~\cite{vaswani_attention_2017}, etc.
MMMPF is not limited to one model but is a general learning task for all time series NN models.
For the loss functions $\{\rho_\tau\}$, we could choose $\tau=\{0.05, 0.50, 0.95\}$ corresponding to the $5\%, 50\%, 95\%$ quantiles.

\begin{algorithm}[!ht]
	\KwIn{Time series model $f_{\mathbf{\theta}}$ with trainable parameters $\mathbf{\theta}$, maximum forecasting horizon $k$, maximum history length $T$, loss functions $\{\rho_\tau\}$}
	\KwData{Time series dataset $S=\{\mathbf{z_i}\}=\{(\mathbf{x_i}, \mathbf{y_i})\}$, where $i$ represents the $i$-th time step, $\mathbf{x_i}$ are the predictor variables, $\mathbf{y_i}$ are the forecast variables}
	Preprocessing dataset with a sliding window of length $(T+k+1)$ to $\{\mathbf{z_{t-T}},...,\mathbf{z_{t-1}},\mathbf{z_{t}},\mathbf{z_{t+1}},...,\mathbf{z_{t+k}}\}$ sequences, where $\mathbf{z_{t}}$ is the sample at current step\;
	Initializing model parameters $\mathbf{\theta}$\;
	\While{not at end of training epochs}{
		\While{not at the end of all mini-batches}{
			Randomly choose a batch of sequences\;
			Randomly choose an integer mask length $l_m$ for this current batch, $0<l_m\leq k+1$\;
			\For{each sequence in the mini-batch}{
				Mask last $l_m$ steps of forecast variables $\mathbf{y}$\;
			}
			Feed masked sequences to model $f_{\mathbf{\theta}}$, generate estimations $\hat{\mathbf{y}}$\ using information of predictor variables from both the past and future $\mathbf{x_{t-T}}, ..., \mathbf{x_{t+k}}$, and unmasked forecast variables $\mathbf{y_{t-T}}, ..., \mathbf{y_{t+k-l_m}}$\;
			Calculate losses on masked predictions $\sum_{i=k-l_m}^{i=k}\rho_\tau(\mathbf{y_i},\hat{\mathbf{y_i}})$\ for all quantiles $\tau$\;
			Backpropagation, update model parameters $\mathbf{\theta}$\;
		}
	}
	\KwOut{Trained model $f_{\mathbf{\theta}}$}
	\caption{MMMPF Training}
	\label{algo:MMMPF-train}
\end{algorithm}


A major advantage of the MMMPF learning task, which uses all the available information to forecast the variable-length masked variables, is that once it is trained, a base NN model can generate forecasts for any forecast length $l_f$ for $0<l_f \leq k+1$, by simply masking the last $l_f$ steps of the desired forecast variables.
Fundamentally, the self-supervised learning approach should learn a representation of the data by being able to fill in the blanks when some forecast variables are masked.
This leads to the flexibility of MMMPF-trained models during inference: they are not restricted to making fixed-length forecasts.
Instead, the models can generate any forecasts of length from $1$ to the maximum forecast horizon $k$.
This could potentially be useful in some real-world applications, e.g., when an electricity load demand forecast model is trained, it needs to be able to make both short-term forecasts for unit commitment and mid-term forecasts for fuel planning and maintenance planning.
Instead of having multiple models for each horizon, a single MMMPF-trained model could complete all of the tasks.

It should also be noted that some existing time series approaches can concatenate future predictor variables $\mathbf{x_t},...,\mathbf{x_{t+k}}$ into the input vectors and train a model to incorporate both future and past information, but because of the fixed concatenation length they can only generate forecasts of the same length (horizon), unlike MMMPF where the forecast horizon is flexible. 


\section{Experimental Results}~\label{sec:exp}
In this section, we apply Masked Multi-Step Multivariate Probabilistic Forecasting (MMMPF) to the ISO New England electricity demand forecasting dataset, and compare the results with existing forecasting formulations.
For fair comparison, MMMPF, RSF, and DMF all use the same base models.
The task is to forecast electricity demand from 1-day to 60-day ahead, given calendar variables and future weather information, as well as the most recent 30-day demand history.
It should be noted that MMMPF is not limited to only 60-day forecasts.
As a general framework, MMMPF can support any length during training, but in reality the weather forecasts for more than 60-day ahead may not available or accurate, thus we limit the problem to short-to-mid-term forecasting.

\subsection{Dataset and Models}~\label{sec:exp:electricity:dataset}
The ISO New England zonal dataset\footnote{Raw data files can be downloaded at \url{https://www.iso-ne.com/isoexpress/web/reports/load-and-demand/-/tree/zone-info}} includes the demands from year 2011 to 2021 for 8 different zones: Connecticut, Maine, Northeast Massachusetts and Boston, Hew Hampshire, Rhode Island, Southeast Massachusetts, Vermont and West/Central Massachusetts.
Hourly data from 2011 to 2020 are used for training (87,672 samples in total), and 2021 used for testing (8,760 samples in total).

For unit commitment, fuel planning, or maintenance planning, forecasting daily peak electricity demand from 1-day ahead to 60-day ahead is often studied~\cite{hong2016probabilistic}.
To accomplish this daily peak forecasting task, the original hourly dataset is downsampled to daily by taking the maximum values of the day.
Predictor variables include month, date, day of week, dry bulb temperature, and dew point temperatures for each zone.
Forecast variables are the electricity demands for each zone, with a total of 8 variables at each step.
For the time series frameworks (MMMPF, RSF, DMF), they use the same base models and hyperparameters for fair comparisons:
\begin{itemize}
	\item \textbf{LSTM}~\cite{hochreiter_long_1997}: $2$ hidden layers, each with dimension $50$.
	\item \textbf{TCN}~\cite{bai_empirical_2018}: $2$ hidden layers with channel size $50$ each, convolutional kernel size $3$ and stride $1$, dilation factor is $2^i$ where $i$ is the $i$-th layer, and dropout rate $0.2$.
	\item \textbf{Transformer}~\cite{vaswani_attention_2017}: model dimension $64$, feed-forward dimension $256$, number of heads $4$, number of encoder layers $2$, and dropout rate $0.1$. Only encoder is used.
\end{itemize}
For these time series models, we train with an Adam optimizer~\cite{kingma_adam_2015} with a learning rate of $0.001$.
Model batch size is $1000$ and the number of epochs is $1000$. 
We report the mean absolute percentage error (MAPE) for test results in this subsection following common practices in the field of electricity demand forecasting.
For training sets, 80\% of the data were used for training and 20\% for validation.
For categorical variables, an embedding layer with dimension $5$ is used.

For SBF, we trained the following standard ML models:
\begin{itemize}
	\item \textbf{LR-O}: ordinary linear regression.
	\item \textbf{LR-R}: Linear model with ridge regression.
	\item \textbf{LR-L}: Linear model with Lasso regression.
	\item \textbf{SVM-L}: Support vector machine (SVM) linear kernel.
	\item \textbf{SVM-RBF}: SVM w/ Radial Basis Function kernel.
	\item \textbf{GP}: Gaussian Process w/ Matern kernel $\nu=1.5$.
	\item \textbf{DT}: Decision tree (DT) with max depth of 5.
	\item \textbf{RF}: Random forest of 100 DT as above.
	\item \textbf{FCNN}: fully-connected NN with 2 layers of 50 neurons.
\end{itemize}

\subsection{Multi-Step Multivariate Probabilistic Forecasting Results}~\label{sec:exp:electricity:ms}
The results for 1-to-60-day ahead electricity demand forecasting is listed in Table~\ref{table:isone} and grouped by different base models and training methods.
MAPE is averaged over 8 different zones and all forecasting horizons from 1 to 60 days.
It can be seen from the right columns that for this multi-step forecasting problem with known future information, MMMPF significantly outperforms RSF and DMF with exactly the same base model and hyperparameters.
This is because MMMPF directly incorporates known future information in its training processes.
From the left column, several traditional ML models perform relatively poorly compared to MMMPF models, because these models are not time-series models, i.e., they do not use recent history at all.
This results illustrate a simple point: when history information and some future information are known, a ML framework should use both to make better forecasts.
However, standard time series frameworks (RSF \& DMF) fails to incorporate all future information, while non-time series models (DBF) fails to consider the history, leading to worse forecasting results than MMMPF.

\begin{table}[ht]
	\centering
	\caption{Mid-term electricity demand forecasting results. Average MAPE for different forecasts (of horizon 1 day to 60 days) is reported for different methods (lower is better). For all NN-based models MAPE is reported on quantile $\tau=0.5$.}
		\begin{tabular}{c|c|c||c|c}
			\hline
			Base Model & Method & MAPE & SBF Model & MAPE \\
			\hline
			\multirow{3}{*}{LSTM} & MMMPF & \textbf{4.99} & LR-O & 13.91 \\ 
			& RSF & 8.63 & LR-R & 13.99 \\
			& DMF & 19.22 & LR-L & 14.95 \\
			\cline{1-3}
			\multirow{3}{*}{TCN} & MMMPF & \textbf{6.00} & SVM-L & 13.53 \\
			& RSF & 9.62 & SVM-RBF & 5.99\\
			& DMF & 18.67 & GP & 6.22 \\
			\cline{1-3}
			\multirow{3}{*}{Transformer} & MMMPF & \textbf{5.56} & DT & 7.85 \\ 
			& RSF & 7.92 & RF & 7.27 \\
			& DMF & 17.72 & FCNN & 7.59 \\
			\hline
		\end{tabular}
	\label{table:isone}
\end{table}

\begin{figure}[t]
	\centering
	\begin{subfigure}[b]{0.9\linewidth}
		\centering
		\includegraphics[width=1.0\linewidth]{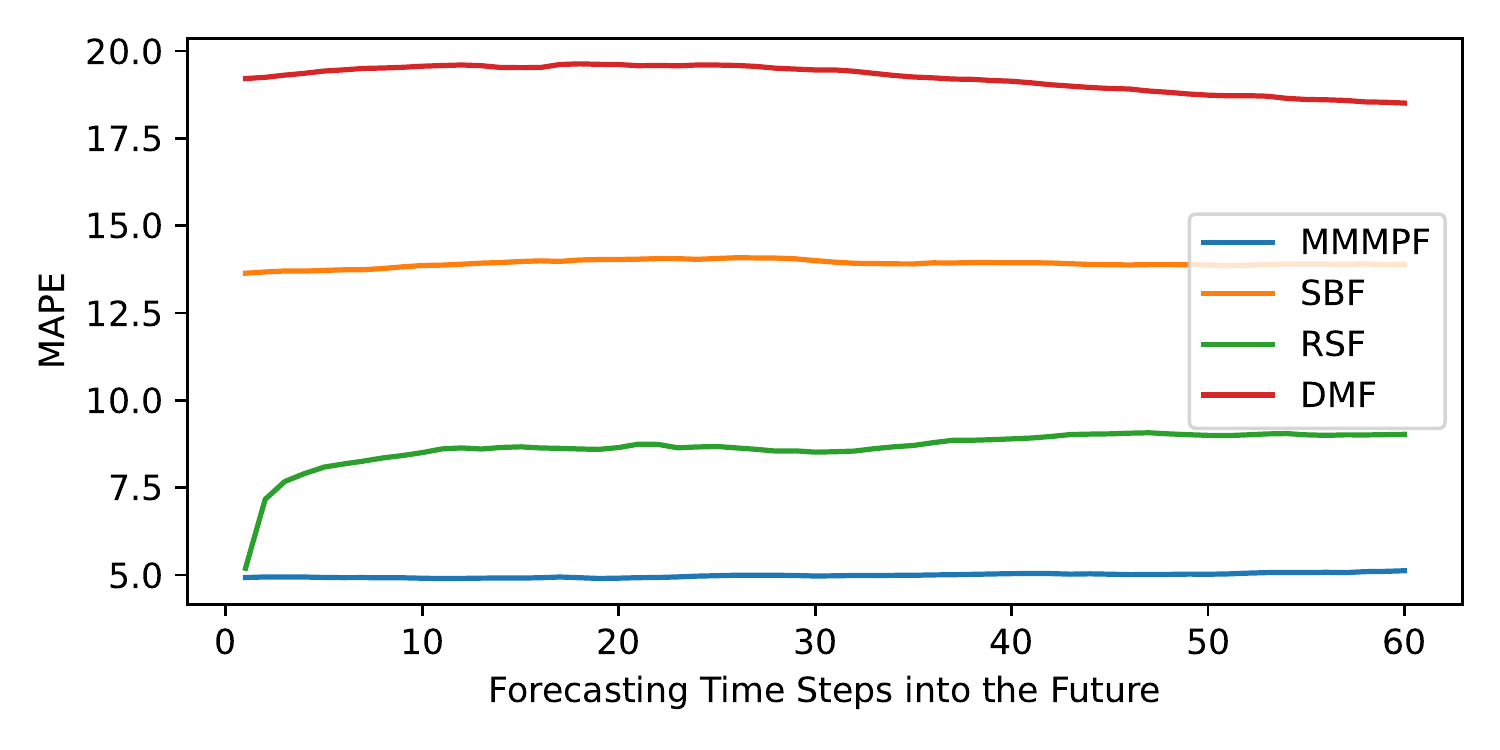}
	\end{subfigure}
	\caption{Daily peak demand forecasting with horizon from 1 to 60 days, average MAPE for 8 zones are shown. MMMPF, RSF, and DMF use LSTM base model, SBF uses LR-O model.}
	\label{fig:elec}
	\vspace{-0.5cm}
\end{figure}

Figure~\ref{fig:elec} shows the MAPE from 1-day to 60-day ahead forecasting.
Because RSF is trained on 1-step forward predictions, its performance worsens significantly as the number of steps increases and predictions are made by using previous predictions.
The performance of DMF is the worst because it needs to directly map the historical data to all 60-step forecasts without using the future information.
Since all 60 steps are being generated simultaneously, the performance does not worsen with increasing forecasting horizon.

\begin{figure}[t]
	\centering
	\begin{subfigure}[b]{\linewidth}
		\centering
		\includegraphics[width=1.0\linewidth]{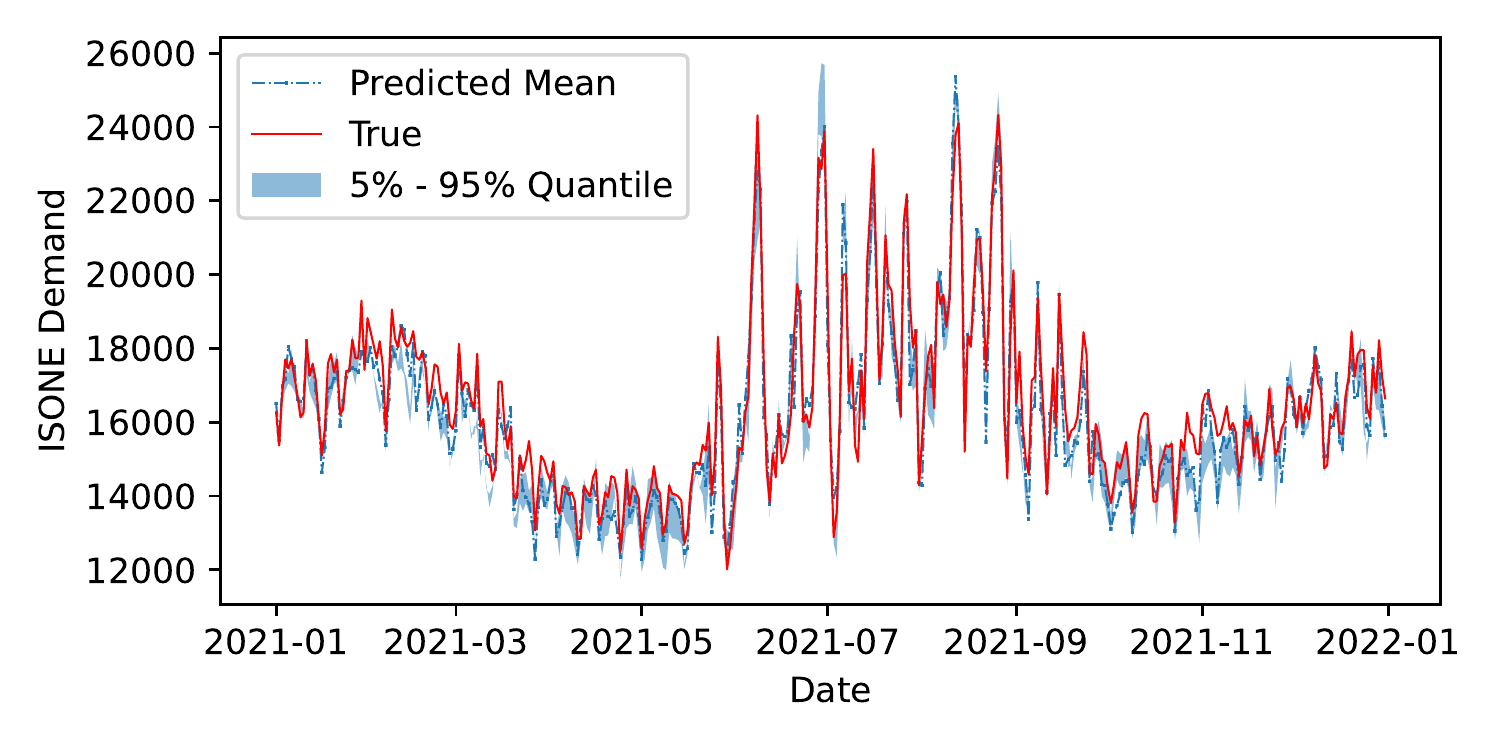}
	\end{subfigure}
	\caption{60-step ahead probabilistic forecasting with MMMPF and LSTM base model.}
	\label{fig:qr}
	\vspace{-0.5cm}
\end{figure}

As probabilistic forecasting becomes increasingly important for the planning and operation of energy systems, the traditional point outputs at each step may not be adequate for many real-world tasks anymore.
Our proposed MMMPF model could generate probabilistic forecasts as shown in Figure~\ref{fig:qr} with models trained by different quantiles $\tau=\{0.05, 0.50, 0.95\}$.

\section{Conclusion}~\label{sec:conclusion}
In this paper, we proposed and experimentally validated Masked Multi-Step Multivariate Probabilistic Forecasting (MMMPF), a new self-supervised learning framework for multi-step time series probabilistic forecasting with known future information. MMMPF will help grid planners and operators obtain multi-step forecasts with uncertainty at several nodes in their network. MMMPF has been experimentally validated on a real-world short-to-mid-term electricity demand forecasting dataset. The main contribution of this paper is to show that MMMPF as a general machine learning task can outperform existing time series forecasting approaches, including recursive methods and direct methods while using the same base model, by incorporating future information; and it can outperform existing non-time series models, by incorporating past information.
Once trained with MMMPF, a time series model can generate any length forecasts below the maximum forecast length during training, as well as different forecasting quantiles. This makes MMMPF an ideal upgrade to any existing deep learning-based multi-step time series forecasting models, and potentially has significant impacts on many other real-world forecasting applications where some future information is available. 
The effect of uncertainty and errors in weather forecasts on the generated demand forecasts with uncertainty will be considered in future work. The calibration of prediction intervals obtained from MMMPF will be explored in future. Other applications, such as renewable energy generation forecasting and electricity spot price forecasting, will be explored in future.


\bibliographystyle{IEEEtran}
\bibliography{main}

\end{document}